\documentclass[a4paper]{article}

\usepackage{INTERSPEECH2021}
\usepackage{cite}
\usepackage{xcolor}
\ninept

\title{Improving the Intent Classification accuracy in Noisy Environment}
\name{Mohamed Nabih Ali $^1 {^,} {^2}$, Alessio Brutti $^2$, Daniele Falavigna $^2$}
\address{
  $^1$ Department of Information Engineering and Computer Science, University of Trento\\
  $^2$ Fondazione Bruno Kessler}
\email{mohamed.nabih@unitn.it, brutti@fbk.eu, falavigna@fbk.eu}

\begin{document}

\maketitle
\begin{abstract}

Intent classification is a fundamental task in the spoken language understanding field that has recently gained the attention of the  scientific community, mainly because of the feasibility of approaching it with end-to-end neural models. In this way, avoiding using intermediate steps, i.e. automatic speech recognition, is possible, thus the propagation of errors due to background noise, spontaneous speech, speaking styles of users, etc. Towards the development of solutions applicable in real scenarios, it is interesting to investigate how environmental noise and related noise reduction techniques to address the intent classification task with end-to-end neural models.

In this paper, we experiment with a noisy version of the fluent speech command data set, combining the intent classifier with a time-domain speech enhancement solution based on Wave-U-Net and considering different training strategies. Experimental results reveal that, for this task, the use of speech enhancement greatly improves the classification accuracy in noisy conditions, in particular when the classification model is trained on enhanced signals. 


\end{abstract}
\noindent\textbf{Index Terms}: Speech Enhancement, Intent Classification, Deep Learning

\section{Introduction}
Spoken Language Understanding (SLU) is a research field that has inspired the interest of scientific communities referring to natural language  processing (NLP) area since many years. Now days,  spoken dialogue interaction, in a natural way,  is possible  with several commercial products, such as the most known personal assistants (Google Home, Amazon Alexa, Siri, Microsof Cortana, ect) and  can be implemented with a set of toolkits, both commercial (e.g. dialog flow \cite{sabharwal2020introduction}) and  open source (e,g, Rasa, Opendial, \cite{Bocklisch2017RasaOS, lison-kennington-2016-opendial}). 

The fundamental  function in SLU systems is the understanding of the intents of the users, which causes the execution of "dialog acts"  aimed to fulfill their requests.
For example, in smart home applications an utterance like "increase the sound" might correspond to an intent represented with the following filled slots: {action: "increase", type:"sound", count:"None", place:"None"}. A survey reporting fundamentals 
of SLU technology can be found in ~\cite{tur2011spoken,gao2019neural}.

The Intent Classification (IC) task is usually accomplished by applying NLP techniques to the output of an automatic speech recognition (ASR) system, in order to produce a semantic interpretation of the input speech. Recently, approaches that perform this task in an end-to-end (E2E) fashion has started to be investigated and produced excellent performance on several field data sets. The  E2E paradigm uses a single neural model to map a spoken input into the corresponding intents, thus optimizing directly the classification metrics and avoiding error propagation caused by
ASR errors. Some interesting models and related results in this direction can be found in the works reported in \cite{lugosch2019speech, haghani2018audio, serdyuk2018towards, qian2017exploring}.



Unfortunately, as in ASR systems, environmental noise deteriorates the quality and and intelligibility of speech signals, resulting in low intent classification accuracy \cite{vary2006digital}. To mitigate the impact of noise, a possible approach consists in training, or adapting, the classification model on the noisy data \cite{yin2015noisy}. This can be done either by collecting application specific data  or through the usage of data augmentation strategies \cite{braun2020data}. However, it has to be considered that acquiring large sets of noisy data is costly and time consuming while, in general, all possible noisy conditions cannot be known a-priori making unfeasible the data augmentation based approach. Therefore, an alternative method is to use a speech enhancement front-end to improve the speech quality . 

In this paper we propose a pipeline that integrates a speech  enhancement front-end based on time-domain approaches with an end-to-end intent classifier implemented with a time-convoluted neural model and investigate the final classification performance for different training strategies. To the best of our knowledge this is the first time in which speech enhancement is applied to an intent classification task within an end-to-end framework. In addition, we also want to point out the great efficiency we have found in using the adopted end-to-end classification network, which implements an architecture analogous to Conv-Tas Net \cite{luo2019conv} and, despite its small size, provides state-of-the-art performance \cite{obuchowski2020transformer, radfar2020end,lugosch2019speech, cho2020speech}.


This rest of this paper is organized as follows: Sect. 2 we briefly review the most of speech enhancement techniques and its drawbacks. In Sect. 3 we presented our proposed approach for both speech enhancement and IC. Our experimental results are showed in Set.4. Lastly, in Sect. 5, conclusion and future directions on speech enhancement and IC are deliberated.  

\begin{figure*}
  \includegraphics[width=\textwidth,height=5cm]{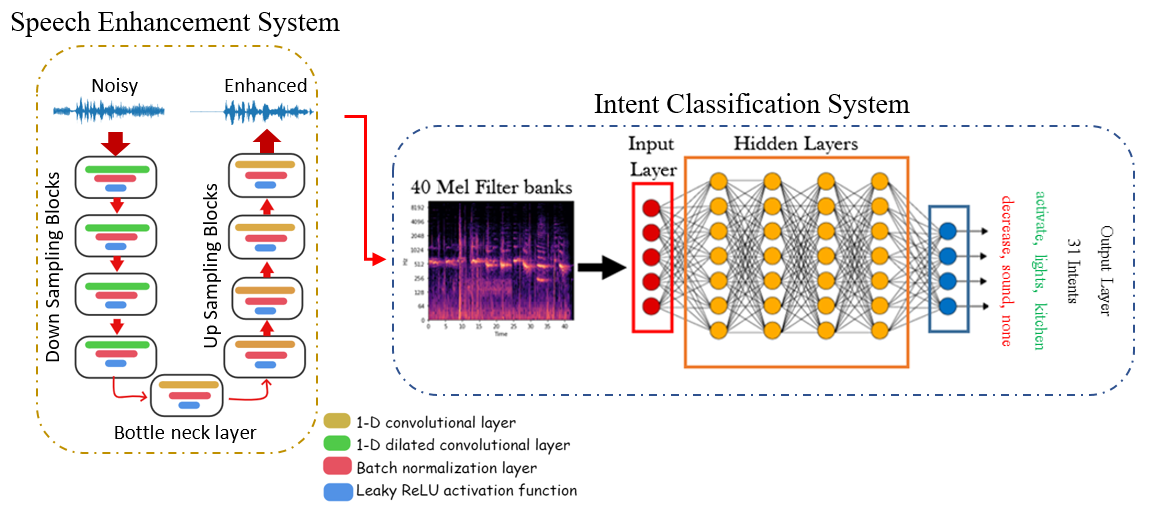}
  \caption{The proposed system of Speech Enhancement and Intent Classification}
  \label{fig:SEcla}
\end{figure*}

\section{Proposed Approach}
In this paper, we tackle the problem of intent classification in noisy environment by employing a speech enhancement front-end to mitigates the noise impact on the speech signals before processing them with the intent classification back-end. Fig.\ref{fig:SEcla} shows the complete pipeline of the proposed approach. This strategy has been investigated several times in literature for ASR related tasks with contrasting results
\cite{astudillo2015integration, ochiai2017does, kinoshita2020improving, sivasankaran2015robust}. In particular, the speech enhancement front-end often introduces distortions unseen by the acoustic models trained on clean speech \cite{narayanan2014investigation}. Despite the drawback of this strategy, it is still useful from a practical perspective, in which it allows modular research on ASR to be robust against noise.

Nevertheless, we believe that, given the peculiarities of the task under analysis, an experimental analysis in this direction is interesting. In particular, in order to correctly classify a user request the system does not need to recognize any single words. Therefore, we expect that the artifacts typically introduced by speech enhancement processing are less relevant. In the following section we describe the enhancement and intent classification components.

\subsection{Speech Enhancement: Wave-U-Net}
\label{Wave-U-Net}
Traditional speech enhancement methods, such as spectral subtraction \cite{berouti1979enhancement} and Wiener filtering \cite{lim1978all}, operates on the short-time Fourier transform (STFT) under stationarity assumption of the noise signals \cite{breithaupt2008novel}. Thus their performance is regrettable in presence of non-stationary noise\cite{loizou2010reasons}. These statistical assumptions are not necessary when employing deep neural supervised solutions capable of providing enhanced STFT representations \cite{xu2014regression} or time-frequency masks \cite{li2017ideal, williamson2015complex, jiang2011performance} starting from a noisy mixture. The main drawbacks of these techniques is that, the estimated signals phase remains noisy, and training DNN to estimate the signals mask often introduce artifacts, which deteriorate the signal quality. Recently, time-domain approaches have been proposed \cite{luo2018tasnet, rethage2018wavenet} which operates directly on the raw wave-forms and provides an enhanced  signal as output. Towards this direction, \cite{pandey2019new} proposed a time-domain model, named U-Net, to map the noisy signals to their corresponding clean versions. This model was later improved towards Wave-U-Net, firstly proposed in \cite{macartney2018improved} for audio source separation, achieving promising results in comparison with other approaches.

The Wave-U-Net model has three main parts: firstly multiple of 1-D fully convolutional down sampling blocks, followed by 1-D convolutional layer called bottleneck layer, finally a series of 1-D fully convolutional layers called upsampling blocks. In this architecture, skip connection is used between each block in the downsampling part to its corresponding upsampling block. More specifically consider, as input to Wave-U-Net, a mixture of noisy signals $y[n]\in [-1, 1]^{L\times C}$, where $C$ is the number of speech channels and $L$ is the number of audio samples. During training, the high level features are computed based on defined time scales using the downsampling blocks. These features are then concatenated with its corresponding local, and high resolution features computed from the upsampling blocks. In case of monaural speech enhancement, the network is trained to estimate the clean speech signals, i.e. $x^1, \dots, x^K$ with $x^k \in [-1, 1]^{L\times C}$.

\subsection{Intent Classification}

The IC task aims to recognize the encoded intents in a given spoken utterance \cite{firdaus2020deep}. This task is a fundamental part of human-machine dialogue management, where the intentions of a user has to be mapped into corresponding actions. This task is usually  implemented by processing the outputs of  ASR systems with NLP tools, while the dialogue policy is learned by a Partially Observable Markov Decision Process \cite{young2013pomdp} and optimized using Reinforcement Learning approaches, also employing  deep neural networks \cite{firdaus2020deep}.

Recently the scientific community is proposing neural end-to-end architectures, where the {\em speech-to-intents} mapping is carried out applying directly the network to the spoken input,  without employing any ASR systems. These approaches have been proved to be effective both on large data sets, such as Google Home \cite{li2017acoustic, haghani2018audio}, and on a smaller data set, such as the Fluent Speech Command~\cite{lugosch2019speech}. As reported in \cite{lugosch2019speech}, the reason for this is manifold: {\em a)} E2E models avoids using either useless information or information contaminated by errors in the ASR output;  {\em b)} it learns directly the metric used in the evaluation phase; and {\em c)} it can take advantage from supra-segmental information contained in the speech signal to process.
In a way similar to what happens for both  ASR and computer vision, the use of pre-trained models has been exploited in several works. In \cite{lugosch2019speech} a model pre-trained on ASR targets is exploited for IC on the Fluent Speech dataset. The work reported in  \cite{cho2020speech} proposes to use knowledge distillation from a big pre-trained model for training and E2E SLU model over Fluent Speech Command. Both papers mentioned above reports very high accuracy (above 98\%) on the IC task. Other works dealing with SLU are  reported in \cite{serdyuk2018towards} (this work doesn't employ any pre-trained model), in \cite{qian2017exploring}, where an auto-encoder is used to initialize a SLU model and \cite{radfar2020end, obuchowski2020transformer} where a transformer model is used to predict the: variable-length domain, intent, and slots vectors conveyed in a speech signal, in an E2E fashion.

In this work we propose an E2E multi-class architecture to directly map  each utterance of the Fluent Speech Command data set  into the corresponding  31 possible intents (as previously seen an input utterance can contain more than one intent). The proposed E2E model is based on  Conv-TAS, a neural architecture  introduced for time audio separation \cite{luo2019conv}. The model receives the outputs of 40-Mel filter banks as an input, computed using window size of 20ms, at 10ms step. The neural architecture consists of a normalization layer followed by 1-D convolutional layer to map the 40-Mel feature into bottleneck feature with 64 channels, followed by 2 blocks each block has 5 residual blocks consists of 1-D dilated convolutional layer, while skip connection is used for the same purpose mentioned in section \ref{Wave-U-Net}. 

Moreover, for this work 
we trained the model over different data sets derived from Fluent Speech Command: clean speech, noisy speech, enhanced speech and some mixture of the three (see section~\ref{sec:experiment} for the details). Finally, contrastive evaluations have been carried out  (clean vs. clean, clean vs. noisy, noisy vs. clean, etc) and the related results are reported.


\begin{table*}[!ht]
\begin{center}
\caption{Enhancement performance of Wave-U-Net on noisy FSC dataset}

\begin{tabular}{|c|c|c|c|c|c|c|}
\hline
 Model& PESQ & STOI & fwSNRseg & Cbak & Csig & Covl\\
\hline
Unproc.& 1.47 & 0.63 & 5.58& 1.61 & 2.10 & 1.70\\
\hline
\hline
Wave-U-Net FSC & 1.78 & 0.70 & 9.56 & 1.98 & 3.14 & 2.57\\
\hline
\end{tabular}
\label{tab: TABLEI.}
\end{center}
\end{table*}

\section{Experimental Results}
\label{sec:experiment}
\subsection{Dataset}
\label{Dataset}
For our experimental analysis we consider the  Fluent Speech  Commands  (FSC) dataset~\cite{lugosch2019speech}. FSC includes 30,043 English utterances, recorded as 16 kHz single channel audio files, spoken by 97 native and non native speakers which interacts with voice-enabled appliances or smart devices. Overall, the dataset provides 248 different utterances that are mapped in 31 different intents. Each intent consists of three items: {action, object, and location}. For example, "increase heat in the kitchen" is categorized as: {action: "increase", object:"heat", location:"kitchen"}. Totally, 6 different actions, 14 objects and 4 locations are available, and the combination of these slots provides the utterance intent. We use the official splits as described in \cite{lugosch2019speech}: 23132 utterances for training, 3118 for validation and 3793 for test. To train the speech enhancement model we use a subset of the dataset, considering 18 speakers for a total of 6209 utterances as training set and 3 speakers, resulting in 950 utterances, as validation set. The enhancement performance is measure on the whole test set.    

To emulate realistic application scenarios where environmental noise may affect the intent classification accuracy, we contaminated the FSC, creating an equivalent noisy version, by superimposing 8 different types of noise out of the 25 noises available in the Microsoft Scalable Noisy Speech Dataset (MS-SNSD) \cite{reddy2019scalable}, namely: Air conditioner, Airport Announcement, Cafeteria, Copy Machine, Kitchen, Restaurant, Shutting Door, and Typing noise, resulting 8 noise files .The FSC dataset is contaminated using the ``maracas'' library available in \cite{JFSantos}. For each clean utterance, a noise signals is randomly selected from those available and superimposed with an SNR randomly selected from 6 possible values: -20dB, -15dB, -5dB, 5dB, 15dB and 20dB. Note that the resulting noisy dataset includes a variety of very different conditions in terms of type and amount of noise.

\subsection{Training Parameters}
The speech enhancement network is designed to process fixed-length input signals with sample length 16834. The model is trained using the mean square error (MSE) loss with the ADAM optimizer. The learning rate is 0.001, decay rates are $\beta1 = 0.9$ and $\beta2 = 0.999$ and the batch size 10. Every 50 epochs the model is evaluated in terms of  intelligibility scores on the validation dataset with patience set to 3. 

For the intent classifier, the model is trained for 100 epochs, with learning rate 0.001, ADAM optimizer, and batch size 100. As the intent classification is a multi-class classification problem the Cross-Entropy loss function is applied to estimate the probability of each class.

\subsection{Speech Enhancement Results}
Although our final goal is to improve the classification accuracy, we also evaluate the performance of the enhancement component. Beside the traditional speech quality metrics, namely PESQ \cite{itu862} and STOI \cite{taal2010short}, we consider also Mean Opinion Score (MOS) (CSIG, CBAK, COVL) to predict the signal distortion, the intrusiveness of background noise, and the overall effect respectively \cite{hu2007evaluation}. We also consider the frequency-weighted segmental SNR (fwSNRseg), computed as \cite{hu2007evaluation}.

\begin{figure}
  \includegraphics[width=8.5cm,height=5cm]{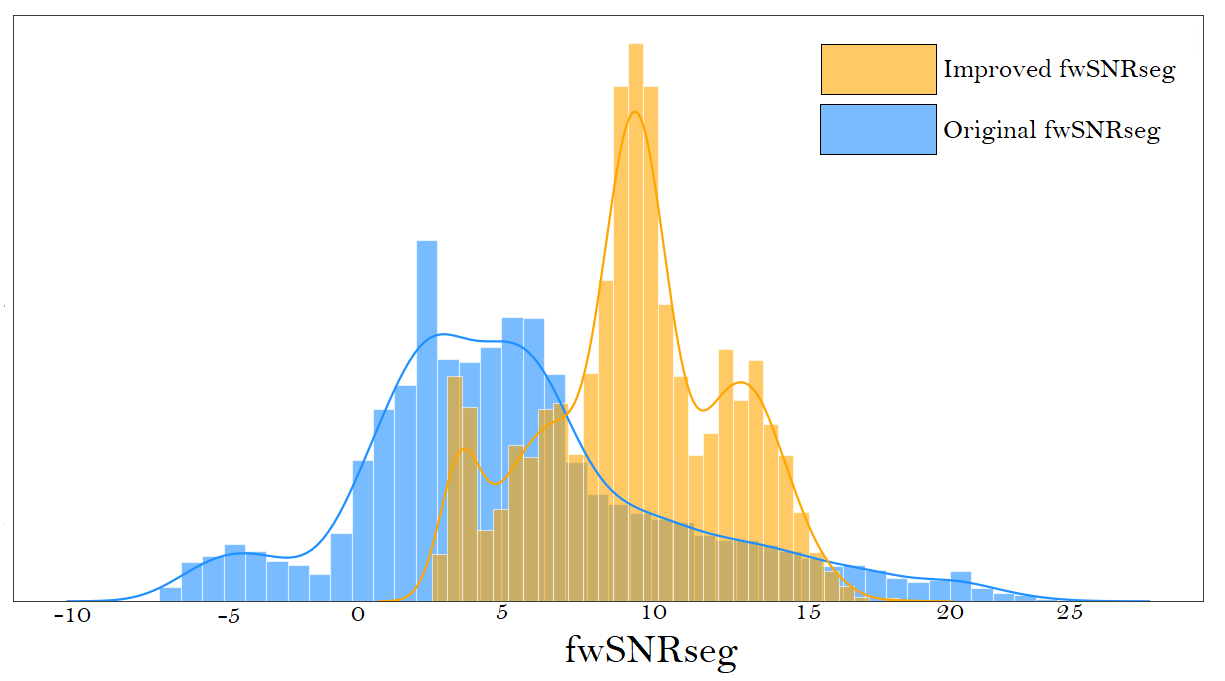}
  \caption{fwSNRseg distribution before and after applying the speech enhancement front-end.}
  \label{fig:DIS}
\end{figure}

Tables \ref{tab: TABLEI.} reports the enhancement performance in terms of the quality and intelligibility metrics reported above. The front-end module improves both the signals quality and intelligibility, measured in-terms of PESQ and STOI metrics, with respect to the unprocessed signals  with scores 1.78 and 0.7 respectively. Moreover, the enhancement substantially improves the MOS scores, considerably reducing the speech distortion measured by the CSIG metric, and effectively eliminating, as measure by CBAK. Subsequently, it scores a better trade-off between the two factors COVL. Finally, we plot in Fig. \ref{fig:DIS} the distribution of the fwSNRseg metric on the test set before and after enhancement. 
In line with the numerical results, the fwSNRseg is improved avoiding negative values hence reducing the impact of noise on the speech signals.   


\begin{table}[!ht]
    \centering
    \caption{Intent classification accuracy on FSC trained and evaluated on different versions of FSC dataset}
    \begin{tabular}{|c|c|c|c|}
    \hline
    Training& \multicolumn{3}{|c|}{Evaluation Data}\\
        \cline{2-4}
        & Clean & Noisy & Enh-FSC \\
        \hline
         Clean      & 98.7\% & 54.3\% & 71.2\% \\
         \hline
         \hline
         Noisy      & - & 70.9\% & 74.4\% \\
         Enh        & - & 54.9\% & 74.8\% \\
         Clean-Noisy & - & 67.8\% & 74.2\% \\
         
         Enh.-Noisy & - & 69.1\% & 74.8\% \\
         \hline
    \end{tabular}
    \label{tab:intent_accuracy}
\end{table}

\subsection{Intent Classification Results}
We evaluate the impact of the speech enhancement component on the intent classification task by considering different training material: clean signals, noisy signals and enhanced signals. Each model is evaluated considering different evaluation data. Table~\ref{tab:intent_accuracy} reports the results. First of all, note that our baseline (i.e. clean model evaluated on clean data) achieves 98.7\% classification accuracy which is aligned with the state of the art on this dataset. Still considering the model trained on the clean data (first row of Table~\ref{tab:intent_accuracy}), as expected the performance on noisy test data dramatically deteriorates (54.4\%). However, already in this conditions, the enhancement front-end considerably attenuates the impact of noisy. One important aspect is that the enhancement module has been trained on in-domain noisy material, matching the noisy conditions in evaluation. This leads to the question of what would be the performance if the classifier was trained on the same noisy material. Of course, as reported in the second row of the table, in this case the performance on noisy data considerably improves (from 54.3\% to 70.9\%). However, if speech enhancement is applied, this further improves the performance up to 74.4\%. Note that our noisy model is, actually, trained in a multi condition way, as the noisy material covers a wide range of SNRs. Moreover, in case of extremely negative SNRs the model probably learns nothing useful. Therefore, if the range of SNRs is shifted towards the positive ones by the enhancement front-end the classification accuracy will inevitably benefit. We evaluate the performance of a classification model trained on the enhanced data which leads to a further minor improvement. However, this gain is rather limited and seems to suggest that enhancing the whole training set is not worth the effort.

Finally we consider another set of experiments by combining clean and noisy as well as noisy and enhanced training sets, which, however, does not bring any particular improvement.

\section{Conclusions}
In this paper we describe an experimental analysis on intent classification in noisy environment where a neural speech enhancement front-end based on Wave-U-Net is combined with an end-to-end intent classification scheme. Experiments on the FSC data contaminated with a set of noises from MS-SNSD shows that, contrary to what observed in other speech related classification tasks, de-noising actually is beneficial in terms of final classification accuracy event when models are trained on matched noisy material.

A future direction is to test the proposed approach based on different dataset e.g. (e.g. the ATIS corpus \cite{hemphill-etal-1990-atis}, the Almawave-SLU corpus \cite{bellomaria2019almawave}, the SLURP corpus \cite{bastianelli2020slurp}). Moreover, other approaches can be used and evaluated as front-end speech enhancement e.g. Conv-TAS network proposed in \cite{luo2019conv}. Finally, we aim to investigate on jointly training both modules, in which the speech enhancement process will be guided based on the performance of intent classifier to produce more discriminative enhanced signals.       

\bibliographystyle{IEEEtran}
\bibliography{mybib}

\end{document}